\documentclass[11pt]{article}
\usepackage{graphicx}
\usepackage{amssymb}
\usepackage{amscd}
\usepackage{mathrsfs}
\usepackage{longtable,lscape}
\usepackage{amsthm}
\usepackage{amsfonts}
\usepackage{amsmath}
\usepackage{bbm}
\usepackage{float}
\usepackage{url}
\usepackage{hyperref}

\newcommand{\captionfonts}{\footnotesize}
\makeatletter  
\long\def\@makecaption#1#2{%
  \vskip\abovecaptionskip
  \sbox\@tempboxa{{\captionfonts #1: #2}}%
  \ifdim \wd\@tempboxa >\hsize
    {\captionfonts #1: #2\par}
  \else
    \hbox to\hsize{\hfil\box\@tempboxa\hfil}%
  \fi
  \vskip\belowcaptionskip}
\makeatother 
\begin{document}
\title{Quantum Entanglement in Corpuses of Documents}
\author{Lester Beltran  and Suzette Geriente  \vspace{0.5 cm} \\ 
        \normalsize\itshape
        Center Leo Apostel for Interdisciplinary Studies
        \\ 
        \normalsize\itshape
         Brussels Free University, Krijgskundestraat 33 \\ 
        \normalsize\itshape
         1160 Brussels, Belgium \\
        \normalsize
        E-Mails: \url{lestercc21@yahoo.com},
           \\ \url{sgeriente83@yahoo.com}
              	\\
              }
\date{}
\maketitle
\begin{abstract}
\noindent
We show that data collected from corpuses of documents violate the 
Clauser-Horne-Shimony-Holt version of Bell's inequality (CHSH inequality) 
and 
therefore indicate  
the presence of quantum entanglement in 
their 
structure.  
We obtain this result by considering two concepts and their combination and coincidence operations 
consisting of searches of co-occurrences of exemplars of these concepts 
in 
specific
corpuses of documents. Measuring the frequencies of these co-occurrences and calculating the relative frequencies as approximate probabilities 
entering 
in the CHSH inequality, we obtain manifest violations of the  
latter 
for all considered corpuses of documents. In comparing these violations with 
those analogously obtained in an 
earlier work for the same combined concepts 
in 
psychological coincidence experiments 
with 
human 
participants, also 
violating the CHSH  
inequality, we identify the entanglement as being carried by the meaning connection between the two considered concepts within the combination they form. 
We explain the stronger violation for the corpuses of 
documents, 
as compared to the violation in the psychology 
experiments, 
as being due to the superior meaning domain of the human mind 
and,  
on 
the other 
side, to the latter 
reaching a broader domain of meaning and being 
possibly also actively influenced during the experimentation. 
We mention some of the issues to be analyzed in future work such as the violations of the CHSH inequality being  
larger 
than the `Cirel'son bound' for all of the 
considered 
corpuses of documents.
\end{abstract}

\section{Introduction\label{introduction}}
Quantum entanglement in human language was studied within the Brussels approach \cite{aertssassolidebianchisozzo2016} to quantum cognition \cite{aertsaerts1995,aertsgabora2005a,aertsgabora2005b,aerts2009,pothosbusemeyer2009,khrennikov2010,busemeyerpothosfrancotrueblood2011,busemeyerbruza2012,aertsbroekaertgaborasozzo2013,aertsgaborasozzo2013,kwampleskacbusemeyer2015,dallachiaragiuntininegri2015a,dallachiaragiuntininegri2015b} by means of psychological experiments on human participants 
about the way language is used by them, 
and several aspects of it were researched \cite{aertssozzo2011,aertssozzo2014,aertsetal2018a,aertsetal2018b,aertsarguelles2018}. In the present article we will investigate how quantum entanglement appears in corpuses of documents. 
We will use the same example of 
the 
two concepts {\it Animal} and {\it Acts}, that entangle in the concept combination 
{\it The Animal Acts}, 
which were studied in the above mentioned psychological experiments and for which 
it 
was proved that the Clauser-Horne-Shimony-Holt  \cite{clauser1969} version of Bell's inequality \cite{bell1964,bell1987} 
(CHSH inequality) 
is violated by the relative frequencies of outcomes of the psychological experiments as approximations of the probabilities for these outcomes to occur \cite{aertssozzo2011,aertssozzo2014}. This time however, instead of collecting 
data from the psychological experiments, we will collect 
data from searches of frequencies of appearance of the respective combinations of exemplars in several corpuses of documents. We will show that, as in the case of the psychological experiments, the collected data violate the CHSH inequality,  
which 
hence indicates 
the presence of entanglement, in the used corpuses of documents. 
We will use three corpuses of documents for our investigation, the corpus `Google Books', which can be found available and free to use at \url{https://googlebooks.byu.edu/x.asp}, the corpus `News on Web' (NOW), which is freely available at \url{https://corpus.byu.edu/now/} and the `Corpus of Contemporary American English' (COCA), which is freely available at \url{https://corpus.byu.edu/coca/}. Google Books is the biggest available corpus, with 560 billion words of books ranging over centuries and scanned by Google. Then comes the NOW corpus which 6 billion words of texts from news and periodicals, 
and finally 
COCA has 560 million words of texts of the types of stories. We have 
tested 
the CHSH inequality on all three corpuses of text such that we could identify the consistency of its violation and compare it with the violation encountered in the psychological experiments on human 
participants 
for the same combination of concepts \cite{aertssozzo2011,aertssozzo2014}.

The present work contributes to a further study of the presence of entanglement in human cognition \cite{bruzakittonelsonmcevoy2009,aertssozzo2011,aertssozzo2014,bruzakittorammsitbon2015,gronchistrambini2017,aertsetal2018a,aertsetal2018b,aertsarguelles2018} as studied
within the quantum cognition research programme. However this time we identify entanglement due to the collection of data violating the CHSH inequality in the structure of corpuses of documents, which means that the present result pertains to a domain of research closely related to quantum cognition which investigates the presence of quantum structure in computer science with applications to information retrieval and natural language processing. This domain of research, `quantum structures in computer 
science', 
developed from 2004 onwards quite parallel to quantum cognition to a flourishing research field \cite{vanrijsbergen2004,aertsczachor2004,widdows2004,schmittnurnberger2007,melucci2008,schmittetal2008,bruza2009,coecke2010,piwowarskietal2010,song-frommholz2010,frommholzetal2010,Ingo2011,bucciomeluccisong2011,melucci2015,wangetal2016}. Two European Union funded consortia, `Quantum Contextual Information Access and Retrieval' (QONTEXT) between 2010 and 2013, and `Quantum Information Access and Retrieval Theory' (QUARTZ)\footnote{University of Padova (IT), The Open University (UK), University of Bedfordshire (UK), Vrije Universiteit Brussel (BE), University of Copenhagen (DE), Brandenburg University of Technology Cottbus-Senftenberg (GE), Linn{\ae}us University (SW).} between 2017 and 2020, to which the authors of the present article are connected, have substantially contributed to the development of the field. The research presented in this article is part of a general investigation of identification of quantum structures, such as contextuality, interference, superposition, entanglement, Bose-Einstein and Fermi-Dirac statistics, in the texts of corpuses of documents, for which recently a general framework was proposed \cite{aertsetal2018e}.

We summarize the content of the present paper in the following. In Section \ref{googlebooks}, we analyse in detail the coincidence operation we performed with corpuses of documents and present the results we obtained in the case of Google Books. In Section \ref{nowcoca}, we instead present the empirical data we collected using NOW and COCA as corpuses of documents. In all cases, we find a significant violation of the CHSH inequality which goes beyond the well known `Cirel'son bound' for quantum mechanical measurements \cite{cirelson1980}. The obtained result is compared in Section \ref{psychologicalexperiment} with the violation of the CHSH inequality that was obtained in experiments  
with human participants \cite{aertssozzo2011,aertssozzo2014}. Next, entanglement by considering collocates type of co-occurrences are studied in Section \ref{collocates}. Finally, Section \ref{conclusion} offers some conclusive remarks on the obtained results.

\section{An entangled combination of concepts in Google Books \label{googlebooks}}
The CHSH inequality \cite{clauser1969} 
\begin{eqnarray} \label{chshinequality}
-2 \le E(A',B')+E(A,B')+E(A',B)-E(A,B) \le 2
\end{eqnarray}
is generally considered in quantum physics as a necessary and sufficient conditions for guaranteeing a local and realistic picture of quantum phenomena. Thus, its empirical violation in Bell-type tests is, according to the scientific community, the most spectacular demonstration of the nonlocal character of such quantum phenomena referred to generally as `entanglement'. However, recently, interest has moved towards the identification of quantum effects in cognitive science, in particular in experiments on conceptual combinations, where remarkable violations of this inequality have already been observed \cite{aertssozzo2011,aertssozzo2014,aertsetal2018a,aertsetal2018b,aertsarguelles2018}, as mentioned in Section \ref{introduction}.

Now, (\ref{chshinequality}) is violated in case the  
intermediate term 
$E(A',B')+E(A,B')+E(A',B)-E(A,B)$ of the inequality is smaller than $-2$ 
or bigger than $2$. For the violation obtained in \cite{aertssozzo2011}  
such intermediate term 
is equal to 
$2.42$. We will show that we obtain for all used corpuses of documents violations that are even stronger and will put forward a hypothesis of why this is the case. Before proceeding 
with 
the calculation, we want to explain the content of the CHSH inequality and how we will collect the data on the considered corpuses of text leading to its violation.

The intermediate term of the CHSH inequality is formed by the `expectation values' of    
four  
coincidence 
experiments or operations $e(A,B)$, $e(A,B')$, $e(A',B)$ and $e(A',B')$. 
For example, $e(A,B)$ is the experiment or operation consisting in jointly performing the measurements of concepts $A$ and $B$, and analogously $e(A,B')$, $e(A',B)$ and $e(A',B')$ consist in jointly performing the measurements of respectively concepts $A$ and $B'$, $A'$ and $B$, and $A'$ and $B'$. The 
`expectation values' 
$E(A,B)$, $E(A,B')$, $E(A',B)$ and $E(A',B')$ 
will be calculated from the data gathered by the experiments or operations $e(A,B)$, $e(A,B')$, $e(A',B)$ and $e(A',B')$, respectively, as it will be explained in the following.

Let us first explain what the experiment or operation $e(A,B)$ is. 
Consider for the concept  {\it Animal} the two exemplars {\it Horse} and {\it Bear}, as outcomes, 
and for the concept {\it Acts} the two exemplars {\it Growls} and {\it Whinnies}, as outcomes. 
The four combinations {\it The Horse Growls}, {\it The Horse Whinnies}, {\it The Bear Growls} and {\it The Bear Whinnies}, where each of  them is an exemplar of {\it The Animal Acts}, constitute the four 
outcomes of the experiment or operation $e(A, B)$, where $A$ and $B$ are jointly measured. To obtain the probabilities associated with these four outcomes, we proceed as follows. In the case of the psychological experiment \cite{aertssozzo2011} we 
proposed 
the four possibilities to each one of the participants in the experiment, and 
asked 
them to choose one of the four. The probabilities 
were 
then easily calculated as the 
large number limit of the
relative frequencies of the choices. 

Confronted with 
the 
texts in documents of a corpus, we retrieve with a search on 
them 
the frequencies of appearances of the four strings `horse growls', `horse whinnies', `bear growls' and `bear whinnies', which in the case of Google Books gives us, $0$, $464$, $247$, and $0$, respectively. This 
means 
that in total the four strings appear $0+464+247+0=711$ times. We calculate the relative frequency of appearance then by dividing each number of appearances by this total number of appearances, and determine the 
approximate 
probabilities in this way. This gives us
\begin{eqnarray}
&&P({\rm horse\ growls}) = P(A_1,B_1)= {0 \over 711} \label{GB_HG}\\
&&P({\rm horse\ whinnies}) = P(A_1,B_2)= {464 \over 711} \\
&&P({\rm bear\ growls}) = P(A_2,B_1)= {247 \over 711} \\
&&P({\rm bear\ whinnies}) = P(A_2,B_2)= {0 \over 711} \label{GB_BW}
\end{eqnarray}
The  obtained expectation value $E(A,B)$ is then, 
using (\ref{GB_HG})--(\ref{GB_BW}),
\begin{eqnarray}
E(A,B)=P(A_1,B_1)-P(A_1,B_2)-P(A_2,B_1)+P(A_2,B_2)=
-1
\end{eqnarray}

The idea is that the choice for {\it Animal} which is {\it Horse} is given the value $+1$, while the choice for {\it Animal} which is {\it Bear} is given the value $-1$. Similarly, the choice for {\it Acts} which is {\it Growls} is given the value $+1$, and the choice for {\it Acts} which is {\it Whinnies} is given the value $-1$. Then, combining these values, we obtain that the choice {\it The Horse Growls} is 
associated with the value $+1$, obtained by multiplying the  value $+1$ for {\it Horse} with the value $+1$ for {\it Growls}. Similarly, the choice {\it The Horse Whinnies} is 
$-1$ (multiplying $+1$ with $-1$),  the choice {\it The Bear Growls} is $-1$ (multiplying $-1$ with $+1$),  
and the choice {\it The Bear Whinnies} is $+1$ (multiplying $-1$ with $-1$).  Then, 
$E(A,B)$ is the `expected value' given  by the probabilities $P(A_1,B_1)$, $P(A_1,B_2)$, $P(A_2,B_1)$ and $P(A_2,B_1)$ of each of these values. Hence $E(A,B)=-1$ means that there is a perfect anti-correlation, and indeed, {\it Horse} anti-correlates with {\it Growls} and {\it Bear} anti-correlates with {\it Whinnies}.

To define the three remaining experiments or operations $e(A,B')$, $e(A',B)$ and $e(A',B')$ that are 
needed to calculate the 
intermediate term 
of the CHSH inequality, we consider two different exemplars for {\it Animal} as well as for {\it Acts}, namely {\it Tiger} and {\it Cat} and {\it Snorts} and {\it Meows}. The three experiments or operations are now defined as 
follows: 
$e(A,B')$ is the experiment or operation where the 
previous 
exemplars for {\it Animal}, {\it Horse} and {\it Bear}, are combined with 
new exemplars for {\it Acts}, {\it Snorts} and {\it Meows}, $e(A',B)$ is the experiment or operation where 
new exemplars for {\it Animal}, {\it Tiger} and {\it Cat}, are combined with the 
previous 
exemplars for {\it Acts}, {\it Growls} and {\it Whinnies}, and 
finally 
$e(A',B')$ is the experiment or operation where for both {\it Animal} and {\it Acts} the new exemplars are combined, hence {\it Tiger} and {\it Cat} with {\it Snorts} and {\it Meows}.

So, for the operation $e(A,B')$, the frequencies of appearances of the four strings `horse snorts', `horse meows', `bear snorts' and `bear meows' in Google Books gives us, $202$, $0$, $0$, and $0$, respectively. This makes that in total the four strings appear 202 times and the relative frequencies of appearance determining the probabilities are
\begin{eqnarray}
&&P({\rm horse\ snorts}) = P(A_1,B'_1)= {202 \over 202} \label{GB_HS} \\
&&P({\rm horse\ meows}) = P(A_1,B'_2)= {0 \over 202} \\
&&P({\rm bear\ snorts}) = P(A_2,B'_1)= {0 \over 202} \\
&&P({\rm bear\ meows}) = P(A_2,B'_2)= {0 \over 202} \label{GB_BM}
\end{eqnarray}
The 
expectation value $E(A,B')$ is then, 
using (\ref{GB_HS})--(\ref{GB_BM}),
\begin{eqnarray}
E(A,B')=P(A_1,B'_1)-P(A_1,B'_2)-P(A_2,B'_1)+P(A_2,B'_2)=
1
\end{eqnarray}
For the operation $e(A',B)$ the frequencies of appearances of the four strings `tiger growls', `tiger whinnies', `cat growls' and `cat whinnies' in Google Books gives us, $97$, $0$, $41$, and $0$, respectively. This makes that in total the four strings appear $138$ times and the relative frequencies of appearance determining the probabilities are
\begin{eqnarray}
&&P({\rm tiger\ growls}) = P(A'_1,B_1)= {97 \over 138} \label{GB_TG}\\
&&P({\rm tiger\ whinnies}) = P(A'_1,B_2)= {0 \over 138} \\
&&P({\rm cat\ growls}) = P(A'_2,B_1)= {41 \over 138} \\
&&P({\rm cat\ whinnies}) = P(A'_2,B_2)= {0 \over 138} \label{GB_CW}
\end{eqnarray}
The 
expectation  value $E(A',B)$ is then,  
using (\ref{GB_TG})--(\ref{GB_CW}),
\begin{eqnarray}
E(A',B)=P(A'_1,B_1)-P(A'_1,B_2)-P(A'_2,B_1)+P(A'_2,B_2)=
{56 \over 138}
\end{eqnarray}
For the operation $e(A',B')$, the frequencies of appearances of the four strings `tiger snorts', `tiger meows', `cat snorts' and `cat meows' in Google Books gives us, $0$, $0$, $0$, and $331$, respectively. This makes that in total the four strings appear $202$ times and the relative frequencies of appearance determining the probabilities are
\begin{eqnarray}
&&P({\rm tiger\ snorts}) = P(A'_1,B'_1)= {0 \over 331} \label{GB_TS} \\
&&P({\rm tiger\ meows}) = P(A'_1,B'_2)= {0 \over 331} \\
&&P({\rm cat\ snorts}) = P(A'_2,B'_1)= {0 \over 331} \\
&&P({\rm cat\ meows}) = P(A'_2,B'_2)= {331 \over 331} \label{GB_CM}
\end{eqnarray}
The  
expectation  expectation value $E(A',B')$ is then, 
using (\ref{GB_TS})--(\ref{GB_CM}),
\begin{eqnarray}
E(A',B')=P(A'_1,B'_1)-P(A'_1,B'_2)-P(A'_2,B'_1)+P(A'_2,B'_2)=
=1
\end{eqnarray}
We have now all 
the requested data 
to calculate the  
intermediate term 
of the CHSH inequality, and this gives
\begin{eqnarray}
E(A',B')+E(A,B')+E(A',B)-E(A,B)=1+1+{56 \over 138}+1=3+{56 \over 138}=3.41
\end{eqnarray}
Hence, 
the CHSH inequality (\ref{chshinequality}) is manifestly violated, and it is more strongly violated than in the case of the psychological 
experiments 
\cite{aertssozzo2011,aertssozzo2014}, the 
intermediate term 
being equal to $3.41$, while the 
intermediate term 
in the case of the psychological experiments 
was 
$2.42$. We also note that the violation we identified for the Google Books corpus is 
greater 
than Cirel'son's bound $2\sqrt{2}\approx 2.83$. It is known that the violation produced by an entangled state with respect to a product measurement within the tensor product Hilbert space cannot be higher than the Cirel'son bound, which means that the entanglement we have identified here will be of a different nature than the 
standard 
quantum one.
It is our intention to 
analyse this phenomenon in detail in a forthcoming article \cite{aertsetal2018c}.

\section{Entanglement in NOW and COCA \label{nowcoca}}
We have collected the 
relative  frequencies for 
the same strings both in the corpus of documents NOW and COCA,
see Section \ref{introduction},
and found the following results.

Let us first consider NOW. For the operation $e(A,B)$, for the frequencies of appearance of the four strings `horse growls', `horse whinnies', `bear growls' and `bear whinnies' we  
found 
$0$, $2$, $6$, and $0$, respectively. This 
means 
that in total the four strings appear $8$ times and the relative frequencies of appearance determining the probabilities are
\begin{eqnarray}
&&P({\rm horse\ growls}) = P(A_1,B_1)= {0 \over 8} \label{NOW_HG} \\
&&P({\rm horse\ whinnies}) = P(A_1,B_2)= {2 \over 8} \\
&&P({\rm bear\ growls}) = P(A_2,B_1)= {6 \over 8} \\
&&P({\rm bear\ whinnies}) = P(A_2,B_2)= {0 \over 8} \label{NOW_BW}
\end{eqnarray}

The
expectation value $E(A,B)$ is then, 
 using (\ref{NOW_HG})--(\ref{NOW_BW}).
\begin{eqnarray}
E(A,B)=P(A_1,B_1)-P(A_1,B_2)-P(A_2,B_1)+P(A_2,B_2)=
-1
\end{eqnarray}
For the operation $e(A,B')$, the frequencies of appearance of the four strings `horse snorts', `horse meows', `bear snorts' and `bear meows' in NOW give us, $1$, $0$, $1$, and 0, respectively. This makes that in total the four strings appear $2$ times and the relative frequencies of appearance determining the probabilities are
\begin{eqnarray}
&&P({\rm horse\ snorts}) = P(A_1,B'_1)= {1 \over 2} \label{NOW_HS} \\
&&P({\rm horse\ meows}) = P(A_1,B'_2)= {0 \over 2} \\
&&P({\rm bear\ snorts}) = P(A_2,B'_1)= {1 \over 2} \\
&&P({\rm bear\ meows}) = P(A_2,B'_2)= {0 \over 2} \label{NOW_BM}
\end{eqnarray}
The   
expectation  value $E(A,B')$ is then, 
using (\ref{NOW_HS})--(\ref{NOW_BM}),
\begin{eqnarray}
E(A,B')=P(A_1,B'_1)-P(A_1,B'_2)-P(A_2,B'_1)+P(A_2,B'_2)=
0
\end{eqnarray}
For the operation $e(A',B)$ the frequencies of appearance of the four strings `tiger growls', `tiger whinnies', `cat growls' and `cat whinnies' in NOW give us, $4$, $0$, $0$, and $0$, respectively. This makes that in total the four strings appear $4$ times and the relative frequencies of appearance determining the probabilities are
\begin{eqnarray}
&&P({\rm tiger\ growls}) = P(A'_1,B_1)= {4 \over 4}  \label{NOW_TG}\\
&&P({\rm tiger\ whinnies}) = P(A'_1,B_2)= {0 \over 4} \\
&&P({\rm cat\ growls}) = P(A'_2,B_1)= {0 \over 4} \\
&&P({\rm cat\ whinnies}) = P(A'_2,B_2)= {0 \over 4} \label{NOW_CW}
\end{eqnarray}
The   
expectation  value $E(A',B)$ is then, 
 using (\ref{NOW_TG})--(\ref{NOW_CW}),
\begin{eqnarray}
E(A',B)=P(A'_1,B_1)-P(A'_1,B_2)-P(A'_2,B_1)+P(A'_2,B_2)=
1
\end{eqnarray}
For the operation $e(A',B')$, the frequencies of appearance of the four strings `tiger snorts', `tiger meows', `cat snorts' and `cat meows' in NOW give us, $0$, $0$, $0$, and $19$, respectively. This makes that in total the four strings appear $19$ times and the relative frequencies of appearance determining the probabilities are
\begin{eqnarray}
&&P({\rm tiger\ snorts}) = P(A'_1,B'_1)= {0 \over 19} \label{NOW_TS} \\
&&P({\rm tiger\ meows}) = P(A'_1,B'_2)= {0 \over 19} \\
&&P({\rm cat\ snorts}) = P(A'_2,B'_1)= {0 \over 19} \\
&&P({\rm cat\ meows}) = P(A'_1,B'_1)= {19 \over 19} \label{NOW_CM}
\end{eqnarray}
The
 expectation value $E(A',B')$ is then, 
 using (\ref{NOW_TS})--(\ref{NOW_CM}),
\begin{eqnarray}
E(A',B')=P(A'_1,B'_1)-P(A'_1,B'_2)-P(A'_2,B'_1)+P(A'_2,B'_2)=
1
\end{eqnarray}
We have now all 
the requested data 
to calculate the
intermediate term 
of the CHSH inequality, and this gives
\begin{eqnarray}
E(A',B')+E(A,B')+E(A',B)-E(A,B)=1+0+1+1=3
\end{eqnarray}
Hence, 
the CHSH inequality (\ref{chshinequality}) is violated again, more strongly than in the psychological 
experiments, 
and also more strongly than the Cirel'son bound.

Let us now consider COCA. For the operation $e(A,B)$, the frequencies of appearances of the four strings `horse growls', `horse whinnies', `bear growls' and `bear whinnies' give rise to the following frequencies in COCA, $0$, $11$, $0$, and $0$, respectively. This makes that in total the four strings appear $11$ times and the relative frequencies of appearance determining the probabilities are
\begin{eqnarray}
&&P({\rm horse\ growls}) = P(A_1,B_1)= {0 \over 11} \label{COCA_HG} \\
&&P({\rm horse\ whinnies}) = P(A_1,B_2)= {11 \over 11} \\
&&P({\rm bear\ growls}) = P(A_2,B_1)= {0 \over 11} \\
&&P({\rm bear\ whinnies}) = P(A_2,B_2)= {0 \over 11} \label{COCA_BW}
\end{eqnarray}
The
 expectation value $E(A,B)$ is then, 
 using (\ref{COCA_HG})--(\ref{COCA_BW}), 
\begin{eqnarray}
E(A,B)=P(A_1,B_1)-P(A_1,B_2)-P(A_2,B_1)+P(A_2,B_2)=
-1
\end{eqnarray}
For the operation $e(A,B')$, the frequencies of appearance of the four strings `horse snorts', `horse meows', `bear snorts' and `bear meows' in COCA give us, $6$, $0$, $0$, and $0$, respectively. This makes that in total the four strings appear $6$ times and the relative frequencies of appearance determining the probabilities are
\begin{eqnarray}
&&P({\rm horse\ snorts}) = P(A_1,B'_1)= {6 \over 6} \label{COCA_HS} \\
&&P({\rm horse\ meows}) = P(A_1,B'_2)= {0 \over 6} \\
&&P({\rm bear\ snorts}) = P(A_2,B'_1)= {0 \over 6} \\
&&P({\rm bear\ meows}) = P(A_2,B'_2)= {0 \over 6} \label{COCA_BM}
\end{eqnarray}
The   
 expectation value $E(A,B')$ is then, 
 using (\ref{COCA_HS})--(\ref{COCA_BM}), 
\begin{eqnarray}
E(A,B')=P(A_1,B'_1)-P(A_1,B'_2)-P(A_2,B'_1)+P(A_2,B'_2)=
1
\end{eqnarray}
For the operation $e(A',B)$ the frequencies of appearance of the four strings `tiger growls', `tiger whinnies', `cat growls' and `cat whinnies' in COCA give us, $2$, $0$, $1$, and $0$, respectively. This makes that in total the four strings appear $3$ times and the relative frequencies of appearance determining the probabilities are
\begin{eqnarray}
&&P({\rm tiger\ growls}) = P(A'_1,B_1)= {2 \over 3} \label{COCA_TG} \\
&&P({\rm tiger\ whinnies}) = P(A'_1,B_2)= {0 \over 3} \\
&&P({\rm cat\ growls}) = P(A'_2,B_1)= {1 \over 3} \\
&&P({\rm cat\ whinnies}) = P(A'_2,B_2)= {0 \over 3} \label{COCA_CW}
\end{eqnarray}
The  
 expectation value $E(A',B)$ is then, 
 using (\ref{COCA_TG})--(\ref{COCA_CW}),
\begin{eqnarray}
E(A',B)=P(A'_1,B_1)-P(A'_1,B_2)-P(A'_2,B_1)+P(A'_2,B_2)=
{1 \over 3}
\end{eqnarray}
For the operation $e(A',B')$, the frequencies of appearance of the four strings `tiger snorts', `tiger meows', `cat snorts' and `cat meows' in COCA give us, $0$, $0$, $0$, and $19$, respectively. This makes that in total the four strings appear $19$ times and the relative frequencies of appearance determining the probabilities are
\begin{eqnarray}
&&P({\rm tiger\ snorts}) = P(A'_1,B'_1)= {0 \over 19} \label{COCA_TS} \\
&&P({\rm tiger\ meows}) = P(A'_1,B'_2)= {0 \over 19} \\
&&P({\rm cat\ snorts}) = P(A'_2,B'_1)= {0 \over 19} \\
&&P({\rm cat\ meows}) = P(A'_2,B'_2)= {19 \over 19} \label{COCA_CM}
\end{eqnarray}
The   
 expectation value $E(A',B')$ is then,
 using (\ref{COCA_TS})--(\ref{COCA_CM}),
\begin{eqnarray}
E(A',B')=P(A'_1,B'_1)-P(A'_1,B'_2)-P(A'_2,B'_1)+P(A'_2,B'_2)=
1
\end{eqnarray}
We have now all 
the requested data 
to calculate the 
intermediate term 
of the CHSH inequality, and this gives
\begin{eqnarray}
E(A',B')+E(A,B')+E(A',B)-E(A,B)=1+1+{1 \over 3}+1=3+{1 \over 3}=3.33
\end{eqnarray}
Hence, 
the CHSH inequality (\ref{chshinequality}) is violated again more strongly than in the case of the psychological 
experiments, 
 and more strongly than the Cirel'son bound.

\section{Comparison with the psychological experiments' violation \label{psychologicalexperiment}}
We have found a violation of the CHSH inequality in Google Books, NOW and COCA with values $3.41$, $3$ and $3.33$, respectively, which are all stronger violations of the CHSH inequality than the one we found with the psychological 
experiments 
in \cite{aertssozzo2011}, where the value of the violation was $2.42$. Let us make explicit the probabilities obtained in the 
latter, so 
that we can interpret the difference. 

We present the obtained results taking into account that $81$ individuals participated in the 
experiments. 
For the experiment $e(AB)$, $4$ subjects chose the example {\it The Horse Growls} as a good example
of the combination {\it The Animal Acts}, $51$ respondents chose {\it The Horse Whinnies}, $21$ respondents chose {\it The Bear Growls}, and $5$ respondents chose {\it The Bear Whinnies}. This means that on a totality of $81$ respondents we 
obtained 
portions 
of $4$, $51$, $21$ and $5$ for the different combinations considered. This allows us to calculate the probability for one of the combinations to be chosen. We have, using the symbols 
of 
Sections \ref{googlebooks} and \ref{nowcoca}, 
\begin{eqnarray}
&&P({\rm horse\ growls}) = P(A_1,B_1)= {4 \over 81} \label{PSY_HG} \\
&&P({\rm horse\ whinnies}) = P(A_1,B_2)= {51 \over 81} \\
&&P({\rm bear\ growls}) = P(A_2,B_1)= {21 \over 81} \\
&&P({\rm bear\ whinnies}) = P(A_2,B_2)= {5 \over 81} \label{PSY_BW}
\end{eqnarray}
If 
we insert (\ref{PSY_HG})--(\ref{PSY_BW}) into the expectation value $E(A,B)$, we then get 
\begin{eqnarray}
E(A,B)=P(A_1,B_1)-P(A_1,B_2)-P(A_2,B_1)+P(A_2,B_2)=
-{63 \over 81}
\end{eqnarray}
For the coincidence experiment $e(AB')$, $48$ respondents chose the example {\it The Horse Snorts} as a good example of the combination {\it The Animal Acts}, $2$ respondents chose {\it The Horse Meows}, $24$ respondents chose {\it The Bear Snorts} and $7$ respondents chose {\it The Bear Meows}. This gives 
\begin{eqnarray}
&&P({\rm horse\ snorts}) = P(A_1,B'_1)= {48 \over 81} \label{PSY_HS} \\
&&P({\rm horse\ meows}) = P(A_1,B'_2)= {2 \over 81} \label{horsemeows} \\
&&P({\rm bear\ snorts}) = P(A_2,B'_1)= {24 \over 81} \\
&&P({\rm bear\ meows}) = P(A_2,B'_2)= {7 \over 81} \label{PSY_BM}
\end{eqnarray}
If we insert (\ref{PSY_HS})--(\ref{PSY_BM}) into the expectation value $E(A,B')$, we then get 
\begin{eqnarray}
E(A,B')=P(A_1,B'_1)-P(A_1,B'_2)-P(A_2,B'_1)+P(A_2,B'_2)=
{29 \over 81}
\end{eqnarray}
For the coincidence experiment $e(A'B)$, $63$
respondents chose the example {\it The Tiger Growls} as a good example of the combination {\it The Animal Acts}, $7$ respondents chose {\it The Tiger Whinnies}, $7$ respondents chose {\it The Cat Growls} and $4$ respondents chose {\it The Cat Whinnies}. This gives 
\begin{eqnarray}
&&P({\rm tiger\ growls}) = P(A'_1,B_1)= {63 \over 81} \label{PSY_TG} \\
&&P({\rm tiger\ whinnies}) = P(A'_1,B_2)= {7 \over 81} \\
&&P({\rm cat\ growls}) = P(A'_2,B_1)= {7 \over 81} \\
&&P({\rm cat\ whinnies}) = P(A'_2,B_2)= {4 \over 81} \label{PSY_CW}
\end{eqnarray}
If we insert (\ref{PSY_TG})--(\ref{PSY_CW}) into the expectation value $E(A',B)$, we then get 
\begin{eqnarray}
E(A',B)=P(A'_1,B_1)-P(A'_1,B_2)-P(A'_2,B_1)+P(A'_2,B_2)=
{53 \over 81}
\end{eqnarray}
For the coincidence experiment $e(A'B')$, $12$ respondents chose the example {\it The Tiger Snorts} as a good example of the combination {\it The Animal Acts}, $7$ respondents chose {\it The Tiger Meows}, $7$ respondents chose {\it The Cat Snorts} and $55$ respondents chose {\it The Cat Meows}. This gives
\begin{eqnarray}
&&P({\rm tiger\ snorts}) = P(A'_1,B'_1)= {12 \over 81} \label{PSY_TS} \\
&&P({\rm tiger\ meows}) = P(A'_1,B'_2)= {7 \over 81} \\
&&P({\rm cat\ snorts}) = P(A'_2,B'_1)= {8 \over 81} \\
&&P({\rm cat\ meows}) = P(A'_1,B'_1)= {54 \over 81} \label{PSY_CM}
\end{eqnarray}
If we insert (\ref{PSY_TS})--(\ref{PSY_CM}) into the expectation value $E(A',B')$, we then get 
\begin{eqnarray}
E(A',B')=P(A'_1,B'_1)-P(A'_1,B'_2)-P(A'_2,B'_1)+P(A'_2,B'_2)=
{51 \over 81}
\end{eqnarray}
We have now all 
the requested data 
to calculate the  
intermediate term 
of the CHSH inequality, and this gives 
\begin{eqnarray}
E(A',B')+E(A,B')+E(A',B)-E(A,B)=
{196 \over 81}=2+{34 \over 81}=2.42
\end{eqnarray}
Hence, 
the CHSH inequality (\ref{chshinequality}) is violated, while Cirel'son bound is not.

We can see that all the examples of violation of the CHSH inequality with data collected from 
corpuses
of documents are stronger than the violation that was measured in the psychological experiment. If we compare the probabilities we can notice another 
difference: 
even the less probable choices, such as {\it The Horse Growls}, {\it The Bear Whinnies}, {\it The Horse Meows}, {\it The Bear Meows}, {\it The Tiger Whinnies}, {\it The Cat Whinnies}, {\it The Tiger Meows}, come out all with a higher probability for the psychological 
experiments 
as compared of what their probability values are for the corpuses of documents. Let us compare them, respectively writing the probabilities in the the following 
order: 
first Google Books, then 
NOW, 
then COCA and then the psychological 
experiments. 
We find
\begin{eqnarray}
&&P({\rm horse\ growls})\ {\rm gives}\ (0, 0, 0, {4 \over 81}) \\
&&P({\rm bear\ whinnies})\ {\rm gives}\ (0, 0, 0, {5 \over 81}) \\
&&P({\rm horse\ meows})\ {\rm gives}\ (0, 0, 0, {2 \over 81}) \\
&&P({\rm bear\ meows})\ {\rm gives}\ (0, 0, 0, {7 \over 81}) \\
&&P({\rm tiger\ whinnies})\ {\rm gives}\ (0, 0, 0, {7 \over 81}) \\
&&P({\rm cat\ whinnies})\ {\rm gives}\ (0, 0, 0, {4 \over 81}) \\
&&P({\rm tiger\ meows})\ {\rm gives}\ (0, 0, 0, {7 \over 81})
\end{eqnarray}
 Neither in Google Books, nor in NOW or in COCA any of the uncommon combinations even appear one time, while amongst the $81$ participants in the psychological experiment there are always a limited number that have chosen one of these uncommon combinations as their preferred one. From personal communication with the authors of the article presenting the data of the psychological 
experiments 
\cite{aertssozzo2011}, we know that there was amazement with respect to the relative high frequencies of outcomes of these uncommon preferences and that there was before the experiment the expectation that at least some of them would also be zero like it is the case for the three consulted corpuses of 
documents. 
It was identified that a subgroup of the $81$ participants had mainly given rise to all of the not common choices, and when questioned why, the general response was that they had paid a lot of attention to one specific sentence of the introductory text that was given to every participant in the experiment. Let us copy and paste this introductory text here.

{\it This study has to do with what we have in mind when we use words that refer to categories, and more
specifically `how we think about examples of categories'. Let us illustrate what we mean. Consider the
category `fruit'. Then `orange' and `strawberry' are two examples of this category, but also `fig' and `olive'
are examples of the same category. In each test of the questionnaire you will be asked to pick one of the
examples of a set of given examples for a specific category. And we would like you to pick that example that you find `a good example' of the category. In case there are more than one example which you find a good example, pick then the one you find the best of all the good examples. In case there are two examples which you both find equally good, and hence hesitate which ones to take, just take then the one you slightly prefer, however slight the preference might be. It is mandatory that you always `pick one and only one example', hence in case of doubt, anyhow pick one and only one example. This is necessary for the experiment to succeed. So, one of the tests could be that the category `fruit' is given, and you are asked to pick one of the examples `orange', `strawberry', `fig' or `olive' as a good example, and in case of doubt the best of the ones you doubt about, and in case you cannot decide, pick one anyhow. Let all aspects of yourself play a role in the choice you make, ratio, but also imagination, feeling, emotion, and whatever.}

The sentence that this subgroup had paid much attention to was the last sentence of this introductory text, i.e. `Let all aspects of yourself play a role in the choice you make, ratio, but also imagination, feeling, emotion, and whatever', and so some of them would say that they had chosen `the tiger meows', because that was what they preferred as 
a 
choice in what they would fantasize for the overall scenery in the imagination that the test brought about to them. And of course, even in 
all the books gathered by Google, the fantasy of a `tiger meowing' has little chance to appear.

This
allows  us to put forward 
the following hypothesis. Although we believe that the corpuses of documents are collections of meaning related very sharply to the human mind, certainly if they are interrogated with a 
planned 
set up by human minds such like we have done in this article, they are very shallow still compared to what a human mind itself 
can carry 
as a worldview. So, a first aspect which explains the differences in probabilities of appearance between strings of meaning in the corpuses of documents and these same entities of meaning in  
human minds, 
is the difference in size. Secondly, and perhaps even more important, 
human minds are active entities, 
with the possibility to adapt to the mere context of a questionnaire itself, for example  
the specific 
sentence at the end of the introduction, while the way we can interrogate a corpus of documents is much more limited. We can search for frequencies of appearance of co-occurrent terms, which is what we did to find the violations of the CHSH inequality. The corpus of documents exists independent of what we exactly are looking for with this specific interrogation, while a human mind being questioned interacts with the question and can be directly  
influenced by it.

Of course, much more important than the differences we  
explained 
above between the data gathered form the corpuses of text and the data collected in the psychological 
experiments, one has to consider 
the similarities. In both cases the CHSH inequality is violated structurally in a completely similar way. 
It is 
the meaning connections incorporated in the considered combinations of concepts and the considered combinations of exemplars that are at the origin of the violation, and these meaning connections are present in exactly the same way in the corpuses of documents as in the human minds being tested in the psychological  
experiments.  

\section{Entanglement and Collocates \label{collocates}}

In the foregoing sections we made searches in the respective corpuses of text for strings of letters. What we mean is that if we, for example, searched the element of the corpus that we used to calculate the relative frequencies, this element would be defined as a string of characters. More concretely, a search for the frequency of appearance of
`horse whinnies' 
was a search for the frequency of appearance of the exact string of characters contained in 
`horse whinnies'.  
This is a very sharp way to identify meaning connections, and for the corpuses of texts that we used, a less sharp way of identifying meaning connection is offered by introducing what is called `collocates'. By means of this technique words that appear in each others neighborhood can be spotted.

Let us explain more in detail how such a measure of co-occurrence in neighborhood is technically devised. We have two words, for example 
`horse'
and
`whinnies', 
then one of them will be considered as the center of an interval of words, let us call it the target word, and let us choose it to be 
`horse'.
One can indicate the number of words that the width of an interval with in its center the target word can have, and we choose for our operation that maximum number available in in the COCA, which is 9 words. This means concretely that whenever the second word 
`whinnies' 
is spotted in a search in the interval of 19 words, 9 words to the left of 
`horse' 
 and 9 words to 
its right, 
it will be registered as a co-occurrence of both words
`horse' and  `whinnies'. 

The aim of the use of collocates in our operation is to loosen the strictness of co-occurrence and already allow such a less strict co-occurrence to be counted in case the target word  
`horse' 
and the 
collocate word
`whinnies'  
appear in each others neighborhood. For example, suppose we consider
`cat' 
as the target word and
`meows'  
as the collocate word and take 9 before and 9 after as the spread of the interval of words, then a piece of text such as `But there, underneath, she sees a skinny orange cat. The cat meows. Ivy's heart  
roars', 
will be counted as a co-occurrence -- it is, by the way, a piece of text that really shows up in the COCA corpus of documents when we did our operation.

We will not repeat the whole scheme of the
operations, 
because
they are 
identical to the foregoing ones, except that the strings of characters identifying the co-occurrences are
now 
replaced by the target words and the collocate words giving rise to the co-occurrences. We found the following results.

For the operation $e(A, B)$, the frequencies of appearances of the four {\it Collocate Pairs} 
`horse growls', `horse whinnies', `bear growls' and `bear whinnies' 
in COCA  
give 
us, 0, 12, 3, and 0, respectively. This 
means 
that in total the four {\it Collocate Pairs} appear 15 times and the relative co-occurrence of appearance determining the probabilities are
\begin{eqnarray}
&&P({\rm horse\ growls}) = P(A_1,B_1)= {0 \over 15} \label{COLO_HG} \\
&&P({\rm horse\ whinnies}) = P(A_1,B_2)= {12 \over 15} \\
&&P({\rm bear\ growls}) = P(A_2,B_1)= {3 \over 15} \\
&&P({\rm bear\ whinnies}) = P(A_2,B_2)= {0 \over 15} \label{COLO_BW}
\end{eqnarray}
The 
expectation value $E(A,B)$ is then,
using (\ref{COLO_HG})--(\ref{COLO_BW}),
\begin{eqnarray}
E(A,B)=P(A_1,B_1)-P(A_1,B_2)-P(A_2,B_1)+P(A_2,B_2)=
-1
\end{eqnarray}
	
For the operation $e(A,B')$ the frequencies of appearances of the four {\it Collocate Pairs} 
`horse snorts' , `horse meows' , `bear snorts' and `bear meows' 
in COCA gives us, 12, 0, 0, and 0, respectively.
The 
four {\it Collocate Pairs} appear 11 times and the relative co-occurrence of appearance determining the probabilities are
\begin{eqnarray}
&&P({\rm horse\ snorts}) = P(A_1,B_1)= {12 \over 12} \label{COLO_HS} \\
&&P({\rm horse\ meows}) = P(A_1,B_2)= {0 \over 12}  \\
&&P({\rm bear\ snorts}) = P(A_2,B_1)= {0 \over 12} \\
&&P({\rm bear\ meows}) = P(A_2,B_2)= {0 \over 12} \label{COLO_BM}
\end{eqnarray}
Their expectation value $E(A,B')$ is then, 
using (\ref{COLO_HS})-(\ref{COLO_BM}),
\begin{eqnarray}
E(A,B')=P(A_1,B'_1)-P(A_1,B'_2)-P(A_2,B'_1)+P(A_2,B'_2)=
1
\end{eqnarray}

As for the third operation $e(A', B)$ the frequencies of appearances of the four {\it Collocate Pairs} 
`tiger growls', `tiger whinnies' , `cat growls' and `cat whinnies' 
in COCA 
give us, 4, 0, 6, and 0, respectively. This
means 
that in total the four {\it Collocate Pair}
appear 
10 times and the relative co-occurrence of appearance determining the probabilities are
\begin{eqnarray}
&&P({\rm tiger\ growls}) = P(A'_1,B_1)= {4 \over 10} \label{COLO_TG} \\
&&P({\rm tiger\ whinnies}) = P(A'_1,B_2)= {0 \over 10} \\
&&P({\rm cat\ growls}) = P(A'_2,B_1)= {6 \over 10} \\
&&P({\rm cat\ whinnies}) = P(A'_2,B_2)= {0 \over 10} \label{COLO_CW}
\end{eqnarray}
The 
expectation value $E(A',B)$ is then, 
using (\ref{COLO_TG})--(\ref{COLO_CW}), 
\begin{eqnarray}
E(A',B)=P(A'_1,B_1)-P(A'_1,B_2)-P(A'_2,B_1)+P(A'_2,B_2)=
-0.2
\end{eqnarray}
For the last operation $e(A',B')$, the frequencies of appearances of the four {\it Collocate Pair} `tiger snorts', `tiger meows', `cat snorts' and `cat meows' in COCA gives us, $0$, $0$, $0$, and $37$. This makes that in total the four {\it Collocate Pairs} appear $37$ times and the relative co-occurrence of appearance determining the probabilities are
\begin{eqnarray}
&&P({\rm tiger\ snorts}) = P(A'_1,B'_1)= {0 \over 37} \label{COLO_TS} \\
&&P({\rm tiger\ meows}) = P(A'_1,B'_2)= {0 \over 37} \\
&&P({\rm cat\ snorts}) = P(A'_2,B'_1)= {0 \over 37} \\
&&P({\rm cat\ meows}) = P(A'_2,B'_2)= {37 \over 37} \label{COLO_CM}
\end{eqnarray}
The 
expectation value $E(A',B')$ is then,
using (\ref{COLO_TS})--(\ref{COLO_CM}), 
\begin{eqnarray}
E(A',B')=P(A'_1,B'_1)-P(A'_1,B'_2)-P(A'_2,B'_1)+P(A'_2,B'_2)=
1
\end{eqnarray}
Finally, 
using the formula for 
the CHSH inequality to verify if there is a
violation, we find 
\begin{eqnarray}
E(A'B')+E(AB')+E(A'B)-E(AB)
= 1+1-0.2-(-1)=2.8
\end{eqnarray}
Again we have a violation of
the CHSH inequality. 
But 
if we compare 
it 
with the violation with value 3.33 obtained for the COCA corpus of documents in Section \ref{nowcoca} 
using 
strict strings of characters identifying co-occurrences, the violation is 
minor and 
closer to the violation we obtained for the psychological
experiments. 

\section{Conclusion \label{conclusion}}
We have shown that data we collected from three corpuses of text, Google Books, NOW and COCA, violate the
CHSH 
version \cite{clauser1969} of Bell's inequality \cite{bell1964,bell1987},
which indicates
the presence of entanglement in the combination of the two concepts {\it Animal} and {\it Acts} into the sentence {\it The Animal Acts}. 
More precisely, in 
Sections \ref{googlebooks} and \ref{nowcoca} we have shown the violation collecting 
data of 
coincidence operations on
different combinations of exemplars of {\it Animal Acts} as co-occurrences in the respective corpuses of documents Google Books, NOW and COCA, by using the search  
engines that are 
available on the 
Web 
for these respective corpuses of documents. 
These 
search engines are very reliable which we 
could test 
in different ways -- the measured frequencies are consistent over time and the sentences where the co-occurrences appear can be explicitly consulted -- which means that the statistics that we derived by calculating 
 the relative frequencies of appearance of each co-occurrence, 
as approximations for the probabilities in the CHSH inequality, 
give 
rise to a good approximation of the probabilities which are present as a `meaning structure' in each one of the corpuses of documents. That very similar and comparable results are obtained in the three corpuses of documents, Google Books, NOW and COCA, proves the deep nature of the presence of this probability structure leading to the violation of the CHSH inequality and hence
straightly indicating  
 the presence of quantum entanglement 
 in each of the corpuses of documents. Our interpretation of this violation of the CHSH inequality is that the entanglement revealed by it is carried by the `meaning connection' between {\it Animal} and {\it Acts} in the combination {\it The Animal Acts}. 
 More concretely, it is because the used corpuses of text all are representations in meaning structure of the human mind, due to the texts contained in them being written by humans, that the `meaning connection' between {\it Animal} and {\it Acts} is engraved in these corpuses of documents. Still more concretely, more often a co-occurrence between {\it Horse} and {\it Whinnies} will appear than a co-occurrence between {\it Horse} and {\it Growls}, simply because the meaning contained in {\it The Animal Acts} makes this be the case for humans living in a world where horses will rather whinny than they will growl.

In Section \ref{psychologicalexperiment} we have compared the violations we obtained in Sections \ref{googlebooks} and \ref{nowcoca} with the violation we obtained for the same combination of concepts {\it The Animal Acts} by means of data collected in a psychological 
experiments 
with human
participants 
\cite{aertssozzo2011,aertssozzo2014} and we found a great similarity between the violations in the different corpuses of documents and the violation in the psychological
experiments. 
This is another confirmation of what we expressed already in the forgoing paragraph, namely that the violation originates in the presence of a meaning connection between {\it Animal} and {\it Acts} in the sentence {\it The Animal Acts}. The violations in the three corpuses of texts are stronger than the violation in the psychological
experiments, 
and we
observed 
that this greater strength of violation
is due to the human participants making statistically 
non-zero 
 some of the very uncommon combinations, such as {\it The Horse Meows}, combinations that all give rise 
instead 
to zero probability in the three corpuses of documents. This is an interesting observation, and we put forward a specific hypothesis about it in Section \ref{psychologicalexperiment}. The hypothesis is that on the one hand the human mind is an active entity much vaster than any of the corpuses of documents, and in this sense it is not strange that {\it Horse} and {\it Meows} have zero co-occurrence in all three corpuses of documents.
Despite 
the enormous amount of stories and books contained in them, it is indeed not obvious that 
a 
sentence 
containing the string `horse meows' 
will occur in even one of them. On the other hand, also for the psychological 
experiments 
we would
easily imagine 
people choosing as their preferred combination {\it The Horse Meows}, if also {\it The Horse Snorts} or {\it The Bear Snorts} are possible choices. Even so, and we can check it in  
(\ref{horsemeows}), 
2 people of the 81 that participated in the psychological
experiments 
preferred {\it The Horse Meows} to the other three possible 
choices. 

In section \ref{psychologicalexperiment}, 
we observed that a specific sentence used as an introduction to the series of experiments 
likely induced a small subgroup of 
them  
to answer to the questions in a very imaginative way, preferring 
to 
imagine a horse 
meowing 
than the boring alternative of a horse 
(or a bear)  
just snorting. Probably some of this little subgroup preferred the bear to meow rather than the horse. Anyhow, also this can be seen as part of our hypothesis, namely that the human mind is 
an 
active and creative entity, being influenced by all little details even 
in 
the way the 
experiments are 
explained. 
Obviously, corpuses of documents also
 contain the richness of the human mind, but in a collapsed and frozen way, no longer to be influenced by the way a search is made. Except of course if the search itself is contextual, but that is definitely not the case for the simple
straightforward 
search engines offered for use on the 
Web 
and connected to the corpuses of documents Google Books, Now and COCA.

In Section \ref{collocates} we have partly tested the hypothesis mentioned in the foregoing paragraph. Indeed, we have redone the
operations for the 
{\it The Animal Acts}
combination 
with the corpus of documents COCA, this time however making use of a more fuzzy search system referred to as `collocates'. Instead of indicating a co-occurrence for {\it Horse} and {\it Whinnies} whenever the string 
`horse whinnies' 
appears in a sentence of the documents contained in COCA, with the collocate search a co-occurrence is registered whenever
the word `whinnies'  
appears in an interval of 9 words before 
or after the word `horse'. 
This introduced fuzziness on the part of the search system moves the corpus of documents COCA closer to the human mind, and this is confirmed within the context of our hypothesis by the CHSH 
inequality being less strongly violated in comparison to searches using strict co-occurences.   

We conclude this article with a remark. We have not investigated here the quantum models in complex Hilbert space that can be constructed to represent the collected data, following the procedures in \cite{aertssozzo2014} and 
showing that quantum entanglement
can be considered to be 
present in the operations we performed on corpuses of documents. We plan to deliver this task in  a forthcoming article \cite{aertsetal2018d}. Here, we limit ourselves to mention that it will turn out that these Hilbert space models 
will show 
that entanglement is 
present 
not only in the state of the considered concepts, but also in the measurements and the evolutions. It is not a very well known fact but, if entanglement is
present not only in the states, but as well as in the measurements and evolutions, then  
the Cirel'son's bound can be
exceeded and 
the violation of CHSH inequalities can even 
reach its 
maximum value of 4. 
This clarifies why the breaking of Cirel'son's bound for the entanglement we identified for all the considered corpuses of text is not incompatible with 
a 
quantum mechanical modeling, something we will investigate in 
more 
detail in a second
planned 
article \cite{aertsetal2018c}.

\end{document}